\documentclass[10pt,twocolumn,letterpaper]{article}

\usepackage{wacv}
\usepackage{times}
\usepackage{epsfig}
\usepackage{graphicx}
\usepackage{amsmath}
\usepackage{amssymb}

\usepackage{enumitem}
\usepackage{color} 
\usepackage{cite}
\usepackage{resizegather}
\usepackage[dvipsnames,table,xcdraw]{xcolor}
\usepackage{soul} 
\usepackage{bm}

\usepackage[dvipsnames]{xcolor}

\def\wacvPaperID{192} %

\wacvfinalcopy %

\ifwacvfinal
\def\assignedStartPage{1} %
\fi

\ifwacvfinal
\usepackage[breaklinks=true,bookmarks=false]{hyperref}
\else
\usepackage[pagebackref=true,breaklinks=true,colorlinks,bookmarks=false]{hyperref}
\fi

\ifwacvfinal
\else
\fi

\begin{document}

\title{Learning to Transfer Visual Effects from Videos to Images}

\author{Christopher Thomas\textsuperscript{1}, Yale Song\textsuperscript{2}, and Adriana Kovashka\textsuperscript{1}\\
University of Pittsburgh\textsuperscript{1}\\
Microsoft Research, Redmond\textsuperscript{2}\\
{\tt\small \{chris, kovashka\}@cs.pitt.edu}\\
{\tt\small yalesong@microsoft.com}
}

\maketitle

\begin{abstract}
We study the problem of animating images by transferring spatio-temporal visual effects (such as melting) from a collection of videos. We tackle two primary challenges in visual effect transfer: 1) how to capture the effect we wish to distill; and 2) how to ensure that only the effect, rather than content or artistic style, is transferred from the source videos to the input image. To address the first challenge, we evaluate five loss functions; the most promising one encourages the generated animations to have similar optical flow and texture motions as the source videos. To address the second challenge, we only allow our model to move existing image pixels from the previous frame, rather than predicting unconstrained pixel values. This forces any visual effects to occur using the input image's pixels, preventing unwanted artistic style or content from the source video from appearing in the output. We evaluate our method in objective and subjective settings, and show interesting qualitative results which demonstrate objects undergoing atypical transformations, such as making a face melt or a deer bloom.

\end{abstract}

\section{Introduction}
\label{sec:intro}

Consumer interest in image manipulation is at an all time high~\cite{cooley_2017}. Platforms like Snapchat and Instagram allow users to take photos and videos, apply filters and effects to them, and share them with friends. These filters have been called the primary draw and ``secret weapon'' of these apps~\cite{pullen_2016}. 
While \emph{artistic style transfer services} \cite{deepart,pikazo,adobe_sensei}
give users control over the artistic style they wish to emulate, for example by allowing users to make their photos look like any painting they choose in terms of color and texture,
most \emph{image manipulation services} (e.g. Snapchat's filters) enable only limited predefined effects that the user can apply, for instance, changing skin tone to a predefined set of tones.
There is clear interest in applying other types of effects to user content. For example, a tutorial showing how to use Adobe After Effects to apply a disintegration effect to a still image has over 750K views.\footnote{https://www.youtube.com/watch?v=K4tPRaxyloE} Unfortunately, replicating this effect requires expertise and costly software, and must be manually redone for each image.

\begin{figure}
\label{fig:concept}
\centering
\includegraphics[width=\linewidth]{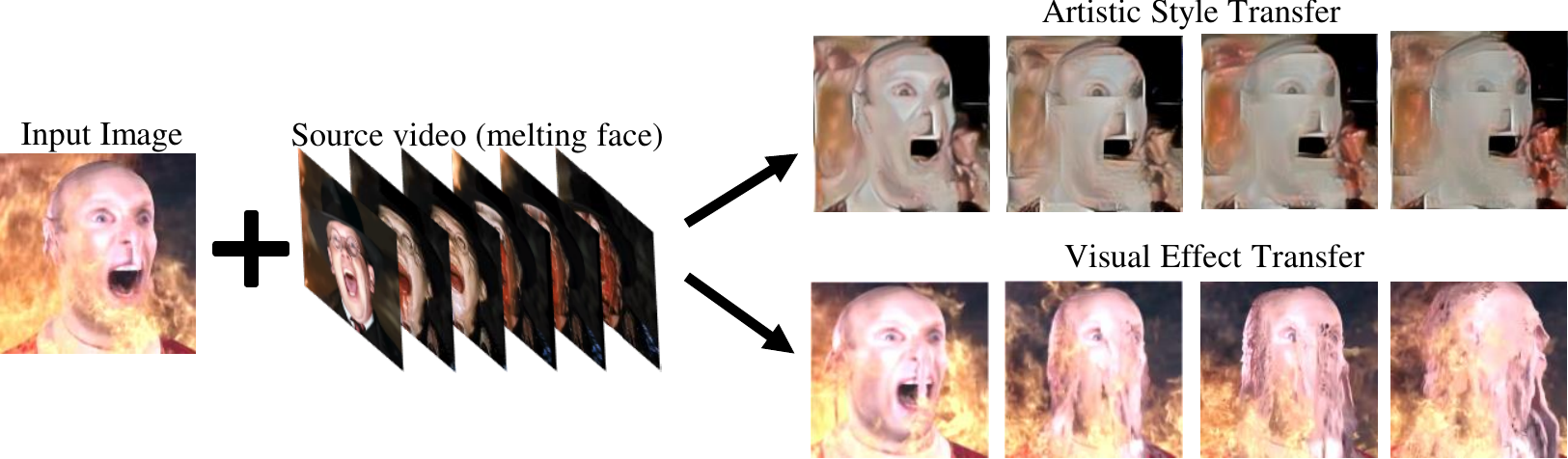}
\caption{\textbf{Visual effect transfer}: Given (1) a user-provided input image, and (2) a source video for a visual effect (e.g. ``melting face''), our goal is to transfer the \emph{effect of the source video} to the target image while preserving the artistic style of the image. In contrast, style transfer methods transform the \emph{style} of the image towards that of the source video, by copying low-level textures and colors and changing the image's appearance. They fail to capture the underlying effect which should deform the object.} 
\end{figure} %

In this paper, we develop methods that allow users to apply \textit{motion-based} visual effects to their own content. 
The user inputs an image and (a) a source video, or (b) the name of a visual effect (e.g. ``melting'' or ``blooming'') whose motion they want to transfer. 
If the latter, we assume that training data for the effect can be found in a dataset (e.g. \cite{zhou2016learning} in our case) or can be downloaded from the internet; how to obtain such training data is \emph{not} the subject of our work. 
We wish to develop an automatic technique for the user to replicate the ``melting'' effect on their own photo, as shown in Fig.~\ref{fig:concept}. 
We refer to a video showing the effect specified by the user as a ``source video,'' and to the user-provided image as the ``input'' or ``target'' image, since it is \emph{to this target} that we want to apply the motion effect. The output of our method is a \emph{new video} which has the motion patterns of the source video and the artistic style of the input image.

Our goal is to distill and transfer the \textit{spatio-temporal visual effect}, while ignoring extraneous signal unrelated to the effect, such as style. By ``style'' we mean the static signature of an image or frame, e.g. the distribution of color intensities. In contrast, the visual ``effects'' we are interested in transferring require modeling the following characteristics: %
\begin{itemize}
\item \textbf{C1: Motion.} The effects we consider all exhibit some form of movement which must be captured from the source video, and then rendered in the style of the target image. For example, Fig.~\ref{fig:concept} requires downward and flowing movement for melting. 
\item \textbf{C2: Object deformation.} Transferring some effects, e.g., baking or blooming, requires objects to structurally deform in certain ways, i.e. bread rises (baking) and leaves may extend from a bud and become pointed (blooming).
\item \textbf{C3: Texture changes.} While we require styles (e.g. color)
in the target image to be preserved, we wish to distill how texture \emph{changes} in source videos. For example, 
during a melting effect, the texture of white vanilla ice-cream might become more smooth; but if we render this effect on rough-textured blue object, we expect the texture change (towards smoothing) to still be rendered using blue pixels (i.e. using the original ``style'' of the input image).  
In Fig.~\ref{fig:concept} 
the result with visual effect transfer contains texture changes of the source video (the face becomes smoother and details are lost as it melts), but do not contain the color patterns of the source video. 
\item \textbf{C4: Temporal dependency.} The effects we consider are non-uniform over time. For instance, 
for the melting effect, at the beginning the object may change slowly, preserving its initial textures, whereas later the object may be entirely liquefied and move differently. 
\end{itemize}

From these characteristics, we see that our visual effects are comprised of temporally registered motion and texture changes. Our goal is to distill these from a series of videos, and then transfer them to a user's photo in order to animate it. This is a particularly challenging problem because it requires learning a good visual representation of video that captures appropriate movement patterns, while discarding appearance information such as colors.

Our problem resembles artistic style transfer \cite{gatys2016image,johnson2016perceptual,Jing2017NeuralST,Gatys_2017_CVPR,gatys2016preserving} but in contrast, we seek to transfer \textit{motion patterns}. 
There have also been efforts to apply artistic style transfer to videos~\cite{ruder2016artistic,Chen_2017_ICCV,huang2017real, gao2020fast} but these methods focus on ensuring that the resulting videos are temporally smooth and coherent. This paper is distinct from these works in that we seek to transfer a spatio-temporal visual effect (rather than artistic style) to a still image, while preserving the image's visual appearance. 
We illustrate this difference in Fig.~\ref{fig:concept}. Artistic style transfer changes the image's colors and textures, and in the case of videos, ensures smoothness in the style-transfered video, but nothing moves. 
In contrast, we seek to transfer a visual effect from a video to an image, without transferring the video's artistic style. The visual effect we wish to distill and replicate is a function of space \emph{and} time, where time is a meaningful factor affecting it rather than a discontinuity that is to be smoothed over. 
We do this by forcing any changes to the image to be performed by \emph{moving existing} image pixels, rather than allowing our method to predict unconstrained pixel values. Thus, any changes occur using the original image pixels, preserving the image's artistic style.

A separate body of work focuses on predicting future frames conditioned on past frames~\cite{Mathieu2015DeepMV, Vondrick_2017_CVPR, vondrick2016generating}. These methods are typically trained on specific domains of video data (e.g.\ a train passing by or a baby's facial movements) \cite{vondrick2016generating,Vondrick_2017_CVPR}; they would therefore fail to generalize when applied to out-of-domain data. Unlike video prediction, we are interested in applying the distilled visual effects on any type of image, which may contain objects that typically would not undergo the visual effect (e.g., a deer blooming). The ability to generalize to out-of-domain data is one of the primary motivations of this work.

One fundamental issue in evaluating visual effect transfer is that there is no single ``correct'' answer~\cite{Jing2017NeuralST}, e.g.~a flower can bloom at different speeds and magnitudes. Following Theis~et~al.~\cite{Theis2015ANO}, we take the view that our results should be evaluated in the context of the ultimate goal of our task: namely, producing results that contain the desired visual effects while remaining visually pleasing to humans. We present four evaluation strategies that measure the effectiveness of our methods both objectively and subjectively. 
Objective metrics measure how reliably motions are transferred and how realistic the produced animations are, as judged by trained classifiers. Subjective metrics measure human opinions about which methods produce the better results, and how appealing these results are. 
We evaluate five choices of loss function to train our framework and find that two of these metrics consistently perform better than the rest. 
We then include qualitative results demonstrating how our method transfers visual effects.%

In summary, we make the following contributions:
\begin{itemize}[itemsep=3pt,topsep=3pt,parsep=0pt,partopsep=0pt]
\item We propose the novel problem of visual effect transfer, and develop a framework to perform such transfer by means of spatio-temporal image transformation. 
\item We test a variety of loss functions that balance realistic visual appearance, motion, and texture changes between an image and source videos. 
\item We present an evaluation strategy for assessing the quality of images animated by our methods.
\end{itemize}

\section{Related Work}
\label{sec:relwork}

\textbf{Style Transfer.} Style transfer attempts to render the content of one image in the artistic style of another. Early methods primarily rely on low-level (and often handcrafted) patch-based texture features~\cite{efros1999texture, hertzmann2001image, kwatra2005texture,shih2014style}.
More recently, impressive results have been achieved using features extracted from pre-trained convolutional neural networks (CNNs) \cite{gatys2016image,johnson2016perceptual,elad2017style,Luan_2017_CVPR,Wang_2017_CVPR}.
Gatys~et~al.~\cite{gatys2016image} showed how style transfer can be formulated as iterative optimization by balancing ``content'' and ``style'' statistics. %
Follow-up works \cite{johnson2016perceptual, ulyanov2016texture, Chen_2017_CVPR,Huang_2017_ICCV,WCT-NIPS-2017, park2019arbitrary, yao2019attention, chen2020explicit} 
improve efficiency by performing style transfer in a single feed-forward pass. \cite{Kim2020DeformableST} integrate geometric constraints into style transfer, but do not learn to transfer spatio-temporal effects as we do.

Na\"ively applying image style transfer to videos results in flickering and discontinuities because of the lack of temporal consistency constraints~\cite{Chen_2017_ICCV}. Recent works integrate optical flow into the loss function and produce smoother video stylization results ~\cite{Anderson2016DeepMovieUO, ruder2016artistic}. Other works train feed-forward networks that perform temporally coherent artistic style transfer on \emph{existing} videos~\cite{Chen_2017_ICCV,huang2017real, gao2020fast}. In contrast, we seek to \emph{generate a video from a still image}. Our output video combines the \textit{spatio-temporal video effect} from another video, while preserving the content and artistic style of the original still image. 
Fig.~\ref{fig:concept} illustrates this difference.

\textbf{Video Prediction. } This paper is related to, but distinct from, work on video prediction~\cite{Mathieu2015DeepMV, vondrick2016generating, Kalchbrenner2017VideoPN, walker2016uncertain}, whose goal is to predict future frames conditioned on past frames. For example, Finn~et~al.~\cite{finn2016unsupervised} explore predicting future frames given the current frame, object states, and actions that a robotic arm might perform. %
A number of similar works predict the future based on models of physical processes~\cite{fragkiadaki_visual_prediction,byravan2017se3, Ehrhardt2017LearningAP, Watters2017VisualIN, warnecke2017convolutional}. They predict future image states by modeling the effect of physical forces %
in the image. 
While our task can be thought of as a type of future prediction where the future temporal changes to be undergone are provided (i.e. the visual effect), \emph{how} these changes should be rendered \emph{in the style of a particular image} is unknown;
we seek qualitatively good transfers, rather than a ``correct'' physical prediction. \cite{Cheng_2020_CVPR} animate still images from time lapse video of the same scene, but do not transfer spatio-temporal texture changes or object deformations as we do.

Other works \cite{chuang2005animating, kwatra2003graphcut, tesfaldet2017} focus on generating moving texture patterns. For example, recent work by Tesfaldet~et~al.~\cite{tesfaldet2017} studies how images can be animated with dynamic textures synthesized from videos (e.g. water stream), while preserving the artistic style of an input image (e.g. a patch from a \emph{painting} of water). 
The resulting animations show movements such as rippling water and flickering flames, but do not change the overall structure or appearance of \emph{objects} over time, as we do in this work. 
\cite{tesfaldet2017} do not learn a \emph{model} of motion, and their method outputs a generated video that mimics the texture dynamics of a single video. %
Unlike these approaches, we seek to distill a visual effect from a collection of source videos and model how the motion and texture changes occur \textit{over time}.

\textbf{Image Transformations. } Previous work has shown that performing pixel transformations rather than pixel prediction simplifies learning and leads to better generation results~\cite{finn2016unsupervised}. Numerous methods for performing transformation have been proposed~\cite{finn2016unsupervised, jaderberg2015spatial,wang2016actions}. Vondrick~et~al.~\cite{Vondrick_2017_CVPR} propose a transformation that interpolates between neighboring pixels using learned coefficients. Finn~et~al.~\cite{finn2016unsupervised} propose to shift the previous frame in multiple directions and then combine the shifted frames using masks. Jaderberg~et~al.~\cite{jaderberg2015spatial} show a differentiable network module capable of performing and learning affine transformations inside neural networks. Our method uses pixel transformations to ensure that only pixels from the image to be transformed appear in the output. 
Similar to these approaches, ours learns to transform pixel values, but under a different objective than prior work.
In our task, transformations are essential in order to to prevent artifacts leaking from the source video into the output frames, highlighting the differences between our task and artistic style transfer.

\textbf{Generative Adversarial Networks (GANs). } Recently GANs have generated much interest in image synthesis~\cite{goodfellow2014generative, Radford2015UnsupervisedRL,zhu2017unpaired} and video prediction~\cite{Vondrick_2017_CVPR, Liang_2017_ICCV}. It has been observed that traditional loss functions, such as mean squared error (MSE), result in blurry outputs, particularly in the presence of uncertainty~\cite{finn2016unsupervised, dosovitskiy2016generating}. GANs do not suffer from this problem because they learn to fit to the mode of the image space, rather than the mean-of-modes like MSE \cite{Vondrick_2017_CVPR}. Thus, GANs tend to produce more natural-looking, crisper results \cite{dosovitskiy2016generating}.
GANs have also been used in the context of style transfer by translating images across domains \cite{Jetchev2016TextureSW, Taigman2016UnsupervisedCI, Zhang2017MultistyleGN, dosovitskiy2016generating, palsson2018generative}. Unlike these, we use \emph{two separate} adversarial discriminators to impose constraints on our transformation model in terms of \emph{both appearance and motion}. 

\textbf{Motion retargeting. } Prior works \cite{monzani2000using,thies2016face2face,rakita2017motion,kim2016retargeting,gleicher1998retargetting,okabe2011creating} have studied the problem of motion retargeting: copying the motion of one object (typically a human, animated character, or face) to another object. These methods typically work by registering points from one object's kinematic structure to another and then enforcing that the ``puppet's'' structure moves in the same way \cite{rakita2017motion}. 
Unlike motion and facial retargeting methods, we are interested in transferring spatio-temporal visual effects from source videos (which may contain many different types of objects undergoing the effect) to still images which may contain arbitrary objects. Moreover, the visual effects we study do not only require motion, but require objects to deform and change appearance. 

\section{Method}
\label{sec:method}
In this section, we present our method for learning and transferring visual effects via pixel transformations. To transfer an effect from a collection of videos onto still images, we need a method that models the effect unfolding over time, and learns to reproduce it on arbitrary images. We propose a convolutional LSTM architecture for this task and explore five different loss functions for training it. 

Given an image $\mathbf{x}$ and a source video clip $\mathbf{s}$ containing the visual effect we wish to capture, our model produces a video clip $\mathbf{y}$ by transforming $\mathbf{x}$ to undergo the visual effect shown in $\mathbf{s}$. This is achieved by loss functions adopted from prior work in the novel context of our proposed pixel transformation framework.
These loss functions consider appearance and motion, and we also include adversarial losses that ensure the realism of the output. We primarily focus on the case $\mathbf{s} = \mathcal{S}$, where $\mathcal{S}$ is the entire training dataset of video clips containing one particular effect to be transferred (e.g.~all melting videos), rather than exactly replicating the patterns found in a single video. 
This allows our model to learn a broader concept of the types of deformations to perform for a particular visual effect. 

Formally, we seek a function $f(\cdot)$ that learns to transform $\mathbf{x}$ to have the same spatio-temporal visual effect as $\mathbf{s}$:
\begin{equation}
f(\mathbf{x}; \mathbf{s}; \bm{\omega}) = \Psi(\mathbf{x},g(\mathbf{x}; \mathbf{s}; \bm{\omega}))
\end{equation}
where $\Psi(\cdot)$ is our pixel transformer with parameters $\bm{\omega}$, described below.

\subsection{Pixel Transformer}
\label{sec:trans}

\begin{figure}[t]
\vspace{-1.25em}
\centering
\includegraphics[width=0.95\linewidth]{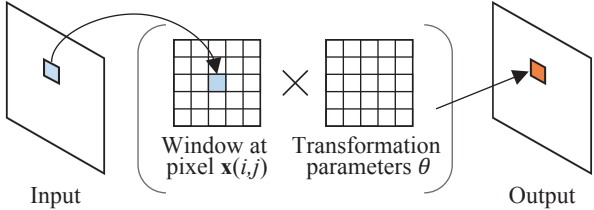}
\caption{Pixel transformation: A window around input pixel $\mathbf{x}(i,j)$ is multiplied by transform parameter $\bm{\theta}$ element-wise and summed to compute an output pixel value. }
\label{fig:pixel_transformer}
\vspace{-1em}
\end{figure}

Our goal is to animate images by transferring the learned visual effect to an image  \textit{without} needing to reconstruct the image's overall appearance. Additionally, we do not want textures or colors from the source videos to be reproduced in the output, since the content in the generated video should match $\textbf{x}$ (e.g. if we are making a face melt, the output video should contain a face, not an ice-cube) and only replicate the visual \emph{effect} shown in $\textbf{s}$.

To accomplish this goal, we define a pixel transformer $\Psi(\bm{x}, \bm{\theta})$ that takes as input an image $\mathbf{x}$ and a transformation kernel parameter $\bm{\theta}$, predicted from a a convolutional LSTM network $g(\mathbf{x}; \mathbf{s}; \bm{\omega})$ with weights $\bm{\omega}$, presented in the next section. 
The kernel $\bm{\theta}$ is defined for every pixel and must be estimated from $\mathbf{s}$, using appropriate losses (Sec.~\ref{sec:losses}).
The transformer produces an output with the pixels in the image moved according to the kernel.
The function $\Psi(\cdot)$ is defined for all pixel locations $(i,j)$ as
\vspace{-0.5em}
\begin{equation}
\label{eq:dna}
\Psi(\mathbf{x},\theta) = \sum_{k \in (-\kappa,\kappa)} \sum_{l \in (-\kappa,\kappa)} \bm{\theta}_{i,j}(k,l) \mathbf{x}(i-k,j-l)
\end{equation}
\vspace{-1em}

\noindent where $\kappa$ is a hyperparameter that determines how far each pixel may possibly move in each direction per output frame; we set $\kappa = 2$. We illustrate our pixel transformer in Fig. \ref{fig:pixel_transformer}.

Our pixel transformer is inspired by Finn~et~al.~\cite{finn2016unsupervised} who proposed convolutional dynamic neural advection (CDNA) that produces multiple shiftings of the entire input image and then combines the shifted images with predicted masks to produce the output. 
While CDNA aims to shift entire images, our goal is to shift pixels, causing objects to deform and undergo non-uniform motion.
We therefore employ a more flexible definition of transformation than CDNA.

\subsection{Network Architecture}

\begin{figure*}[t]
\centering
\includegraphics[width=1\linewidth,height=1.25in]{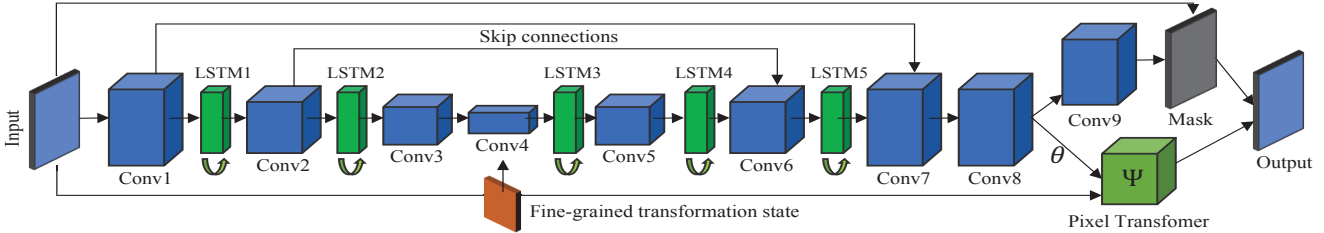}
\vspace{-1.5em}
\caption{The architecture of our network $g(\mathbf{x}; \mathbf{s}; \bm{\omega})$ that predicts the parameter $\bm{\theta}$ for the pixel transformer $\Psi(\cdot)$.}
\label{fig:network_architecture}
\vspace{-1em}
\end{figure*}

Our network $g(\mathbf{x}; \mathbf{s}; \bm{\omega})$ predicts the pixel transformation parameters $\bm{\theta}$ based on an hourglass architecture with convolutional LSTMs, shown in Fig.\ \ref{fig:network_architecture}. It downsamples the input through convolutions and then upsamples the learned representation with deconvolutions. Because it has been reported that deconvolution creates ``checkerboard'' artifacts caused by deconvolution overlap~\cite{odena2016deconvolution}, we replace each deconvolution layer with nearest-neighbor interpolation followed by convolution. LSTMs appear after each intermediate convolution or deconvolution layer to enable the network to maintain state. In our case, this enables the network to learn how the transformation progresses temporally over multiple frames, and how to reproduce it across multiple outputs. Maintaining state allows us to model the temporally-dependent characteristic (\textbf{C4}) of visual effects.

The dataset we use to train our method \cite{zhou2016learning} contains both broad (e.g.~melting) and fine-grained (e.g.~ice-cube melting, butter melting, etc.) annotations. Though we train one model for each broad category, we also provide the fine-grained category annotation to our model as a form of conditioning at training time. This allows our model to learn finer-grained differences in the visual effects exhibited by our videos without being forced to learn an average model, which may smooth over some desirable effects. 

Similar to Finn~et~al.~\cite{finn2016unsupervised}, we predict a background mask to enable pixels from the input frame to be copied directly to the output (for pixels not undergoing a transformation). This enables the network to fill holes caused by moving pixels. We use symmetric padding (zero-padding results in black pixels leaking in).

\subsection{Loss Functions}
\label{sec:losses}
Because $f(\mathbf{x}; \mathbf{s}; \bm{\omega})$ is differentiable, we can train our transformation network end-to-end with our choice of loss function(s). In order to explore the best loss function for our novel motion transfer problem, we implement a number of loss functions from the style transfer and video prediction literature,
which enables easy exploration of loss function combinations.

We implement losses which capture the following components: 1) an \textbf{optical flow} which encourages the output frames to have the same motion patterns as $\mathbf{s}$; 2) a \textbf{motion discriminator} which ensures that the optical flow induced by the generated frame could have come from the real motion distribution learned from $\mathcal{S}$; 3) an \textbf{appearance discriminator} which ensures that the produced frame appearance does not deviate too far from the appearance distribution learned from $\mathcal{S}$; and finally 4) \textbf{style} and \textbf{content} components which encourage the output to have the same appearance patterns as the transformation video $\mathbf{s}$ (e.g.~a melting effect may produce smoother textures over time as the object melts, and this smoothness should be replicated using the pixels of the input image; or a blooming deer might expand in the same way that a blooming flower expands). Note that we only ever use the style component in combination with the other losses, to produce more visually appealing results. The videos output by our framework balance the style of the input image, by virtue of being generated via pixel \emph{transformation}, and the natural look of the visual effect, learned from the source video.
We describe the more promising or unfamiliar losses we explored. 

\textbf{Optical Flow Loss. } We encourage our output video $\mathbf{y}$ to have the same motion representations as the source video $\mathbf{s}$ containing the desired visual effect. This causes $\mathbf{y}$ to exhibit the same motion dynamics as $\mathbf{s}$, addressing characteristics \textbf{C1} and \textbf{C2}, with the added benefit of encouraging temporal smoothness by preventing jarring changes (i.e.\ flickering) and suppressing movements not also in $\mathbf{s}$ (i.e.\ jittering pixels).

Let $\Xi(\mathbf{s}_{i-1}, \mathbf{s}_{i})_l$ represent the activations of layer $l$ of a neural network trained to predict optical flow, on source frames $\mathbf{s}_{i-1}$ and $\mathbf{s}_{i}$, with $1 \leq i < T$, where $T$ is the number of frames of the video containing the visual effect to be copied. Analogously, let $\Xi(\mathbf{y}_{i-1}, \mathbf{y}_{i})_l$ represent the neural activations of layer $l$ on output frame $\mathbf{y}_i$, where $\mathbf{y}_i = \Psi(\mathbf{y}_{i-1}, g(\mathbf{y}_{i-1}; \mathbf{s}_i; \bm{\omega}))$, with $\mathbf{y}_0 = \mathbf{x}$. We denote the shape of $\Xi_l$ by $C_l \times H_l \times W_l$, which represents the number of channels, height, and width of layer $l$'s neural activations. We define $\mathcal{L}_\textrm{flow}(\mathbf{y}_{i-1},\mathbf{y}_i; \mathbf{s}_{i-1}, \mathbf{s}_{i})$ as the normalized difference between $\Xi(\mathbf{s}_{i-1}, \mathbf{s}_{i})_l$ and $\Xi(\mathbf{y}_{i-1}, \mathbf{y}_{i})_l$:
\vspace{-0.05in}
\begin{gather}
\mathcal{L}_\textrm{flow} = \sum_l \frac{1}{C_l H_l W_l} \left\lVert \Xi(\mathbf{y}_{i-1}, \mathbf{y}_{i})_l - \Xi(\mathbf{s}_{i-1}, \mathbf{s}_{i})_l \right\rVert_{2}^2
\label{eq:loss-of}
\end{gather}

\noindent For $\Xi(\cdot, \cdot)$ we use FlowNet2-SD \cite{ilg2017flownet}, with $l \in $ \verb|{conv[1-6]_1,blob[26,31,36,41],|\\
\verb|predict_flow_final}|.

\textbf{Adversarial Motion Loss:} Our previous loss term encourages our network to produce transformations that induce the same optical flow features on $\mathbf{y}$ as $\mathbf{s}$. While this is desirable, the overall motion transformation learned by the model across $\mathcal{S}$ may be unnatural (i.e. the mean of flow modes rather than the mode flow). To address this, we enforce the optical flow of $\mathbf{y}$ to appear as if it were sampled from $\mathcal{S}$'s flow distribution. This discriminator addresses \textbf{C1, C4} by ensuring that the motion (\textbf{C1}) shown occurring over time (\textbf{C4}).
We enforce this with an optical flow discriminator, $\mathcal{D}_\textrm{F}$, trained concurrently with $f(\cdot)$. 

We train $\mathcal{D}_\textrm{F}$ with binary cross-entropy loss. Our predictor seeks to maximize the %
loss, $\mathcal{L}_{\mathcal{D}_\textrm{F}} \left(\Xi\left(I^{(i-1)}, I^{(i)}\right) \right)$, while the discriminator minimizes it:
\begin{gather}
\mathcal{L}_{\mathcal{D}_\textrm{F}} = 
-\left[ d^{(i)} \log \: \mathcal{D}_\textrm{F}\left(\Xi\left(\cdot, \cdot\right)\right) + \right. 
\left. \left(1 - d^{(i)} \right) \log \left(1 - \, \mathcal{D}_\textrm{F}\left( \Xi\left(\cdot,\cdot\right) \right) \right)
\vphantom{d^{(i)} \log \: \mathcal{D}_\textrm{F}\left(\Xi\left(I^{(i-1)}, I^{(i)}\right)\right)} \right]
\label{eq:loss-df}
\end{gather}
\vspace{-0.2in}

\noindent where $\mathcal{D}_\textrm{F}\left(\Xi\left(I^{(i-1)}, I^{(i)}\right)\right)$ is the output of the discriminator on the optical flow map between frames $I^{(i-1)}$ and $I^{(i)}$, both of which are either from $\textbf{s}$ or $\textbf{y}$, and $d^{(i)}$ is a binary label indicating whether the frames come from $\textbf{s}$ or $\textbf{y}$. 

\textbf{Style loss:} In addition to transferring motion, we encourage texture patterns from our source video $\mathbf{s}$ to manifest themselves using the pixels of our input image $\mathbf{x}$; this captures visual effect characteristic \textbf{C3}. 
Let $\Phi_l(\mathbf{s}_i)$ denote the neural activations from layer $l$ of a pretrained VGG-16 \cite{Simonyan2014VeryDC} on visual effect source video frame $\mathbf{s}_i$ (and equivalently for $\mathbf{y}_i$).
Because $\Phi_l(\cdot)$ preserve spatial information and we do not wish to copy the texture patterns at necessarily the same spatial locations, 
we first compute spatially invariant statistics representing the correlations of activations between different neural filters and then compare them, following \cite{gatys2016image}. We do this by computing the Gram matrix $G^{\Phi_l}$, where the matrix entry $G^{\Phi_l}_{a,b}$ represents the inner product between the flattened activations of channels $C_{l_a}$ and $C_{l_b}$ from $\Phi_l$. The Gram matrix is therefore given as
\begin{equation}
G^{\Phi_l}_{a,b} = \frac{1}{C_l H_l W_l} \sum_{h=1}^{H_l} \sum_{w=1}^{W_l} \Phi_{l_{a,h,w}} \Phi_{l_{b,h,w}}
\end{equation}
for all channels, $a,b$ in $C_l$.
Our style loss is given by the normalized difference between the Gram matrices of the output frame $\mathbf{y}_i$ and source frame $\mathbf{s}_i$:
\vspace{-0.05in}
\begin{equation}
\mathcal{L}_\textrm{style}(\mathbf{y}_i, \mathbf{s}_i) = \sum_l \left\lVert  G^{\Phi_l}\left( \mathbf{y}_i \right) - G^{\Phi_l}\left( \mathbf{s}_i \right) \right\rVert_{F}^2
\vspace{-0.3cm}
\label{eq:loss-style} 
\end{equation}
\vspace{-0.15in}

\noindent where $\|\cdot\|_{F}$ is the Frobenius norm. We use VGG-16 layers $l \in $ \verb|{relu[1-5]_1}|.

\textbf{Adversarial Appearance Loss. } While our previous loss functions transfer visual effects, they do not impose any conditions on the quality of the generated frames. This allows unnatural transfers 
to occur and does not penalize blurry outputs. For example, inducing the same optical flow as $\mathbf{s}$ may cause an object to move, but does not impose any constraint on what the pixels should be where the object used to be. 
We also need to ensure that visual effects 
do not affect objects in unnatural ways (\textbf{C2}). 
To address this, we impose a constraint, $\mathcal{L}_{\mathcal{D}_\textrm{A}}\left(I^{(i)}\right)$, that $\mathbf{y}$ should appear to be drawn from the distribution of ground truth source video frames, by concurrently training an appearance discriminator network $\mathcal{D}_\textrm{A}$ that learns to distinguish real frames from transformed frames: 
\begin{gather}
\mathcal{L}_{\mathcal{D}_\textrm{A}} = -\left[ d^{(i)} \log \: \mathcal{D}_\textrm{A}\left(I^{(i)}\right) + \right. 
\left. \left(1 - d^{(i)} \right) \log \left(1 - \, \mathcal{D}_\textrm{A}\left( I^{(i)} \right) \right)
\vphantom{d^{(i)} \log \: \mathcal{D}_\textrm{A}\left(I^{(i)}\right)} \right]
\label{eq:loss-da}
\end{gather}
\vspace{-0.3in}
\subsection{Implementation Details}

\label{subsec:method-implementation}
We train our predictor and discriminator networks end-to-end, but we do not train the networks used only for feature extraction (VGG and FlowNet2). 
We use the Adam optimizer~\cite{Kingma2014AdamAM} with learning rate 1.0e-5. Training our model requires 2 GPUs and a small minibatch size (4) due to memory limitations and takes approximately 2 days. 
We perform teacher forcing \cite{lamb2016professor} for approximately the first 10K iterations, gradually decaying the percentage of ground truth data used in the sequence loss computation according to the schedule $900 / (900 + e^{\frac{\textrm{itr}}{900}})$. We clip the global gradient norm \cite{pascanu2013difficulty} across all predictor layers to be $\leq 0.01$, which we found to help stabilize training. 
For the adversarial losses, we perform one update of the generator and one update of the discriminator every iteration, for 50,000 iterations. 
We add small random noise to the discriminators' input to prevent the discriminators from never being fooled. We use leaky ReLU in our discriminator as suggested by \cite{Radford2015UnsupervisedRL}. 
We modify \cite{finn2016unsupervised}'s codebase to implement our model. Please see the supplementary\footnote{\url{http://www.cs.pitt.edu/~chris/files/2020/motion_supp.zip}} for additional implementation details.

\section{Experimental Evaluation}
\label{sec:results}

\subsection{Evaluation Criteria}
We present four strategies to evaluate effect transfer and discuss their pros/cons.

\textbf{Human Judgments. } 
Following past work on artistic style transfer~\cite{Jing2017NeuralST} and video prediction~\cite{Vondrick_2017_CVPR, vondrick2016generating}, we perform a perceptual study, where we present participants with video clips showing a randomly chosen image 
undergoing a particular transferred effect, all generated from our framework but using training with different losses (Sec.~\ref{sec:baselines}). Participants are then asked to (1) choose the best animation within each set of clips, and (2) for their chosen animation, rate its quality of representing the transformation, on a scale of 1-5, with 5 being best. We recruited 16 non-author participants, each of whom judged 50 sets of clips. We include additional details and the user interface in supp.
We randomize the order of the methods and transformations shown to our participants. %

\textbf{Reconstruction Metrics. } Following prior works~\cite{Mathieu2015DeepMV,Liang_2017_ICCV}, we compute the Mean Squared Error (MSE), Peak Signal to Noise Ratio (PSNR), and Structural Similarity (SSIM) index~\cite{wang2004image}. We provide our models with an image and a source video, and generate an output video using this image as the first frame. Our models may have learned to transfer the visual effect faster or slower than the souce video. To account for this, for all methods, we compare each generated video with multiple versions of the source, each obtained by starting from the same first frame and sampling the same number of frames, but at a different frame rate. We then report the minimum MSE (since lower MSE scores are better) and maximum PSNR and SSIM. Because we are not interested in reconstructing ground truth frames exactly (e.g.~a flower can reasonably bloom in different ways), these metrics are less informative for our task, but do provide some information about the quality of generations. It should be noted, however, that MSE tends to reward blurry predictions~\cite{Mathieu2015DeepMV}.

\textbf{Data Utility. } Recent works have explored training classifiers on synthetic data~\cite{Bousmalis_2017_CVPR, Shrivastava_2017_CVPR}. The better these classifiers perform, the closer the synthetic data is to the target distribution. We train convolutional LSTMs on synthetic data (architecture in supp) to perform visual effect classification. To prevent our classifiers from ``cheating'' by learning to associate objects with effects (e.g.~cookie means baking), we generate training clips showing the same starting frame undergoing each of our effects. Thus, the classifier must learn the spatio-temporal visual effect rather than image appearance. We train one classifier per method and report the accuracy of testing each classifier on the ground truth clips.

\begin{figure*}[t]
\centering
\vspace{-1em}
\includegraphics[width=1\linewidth]{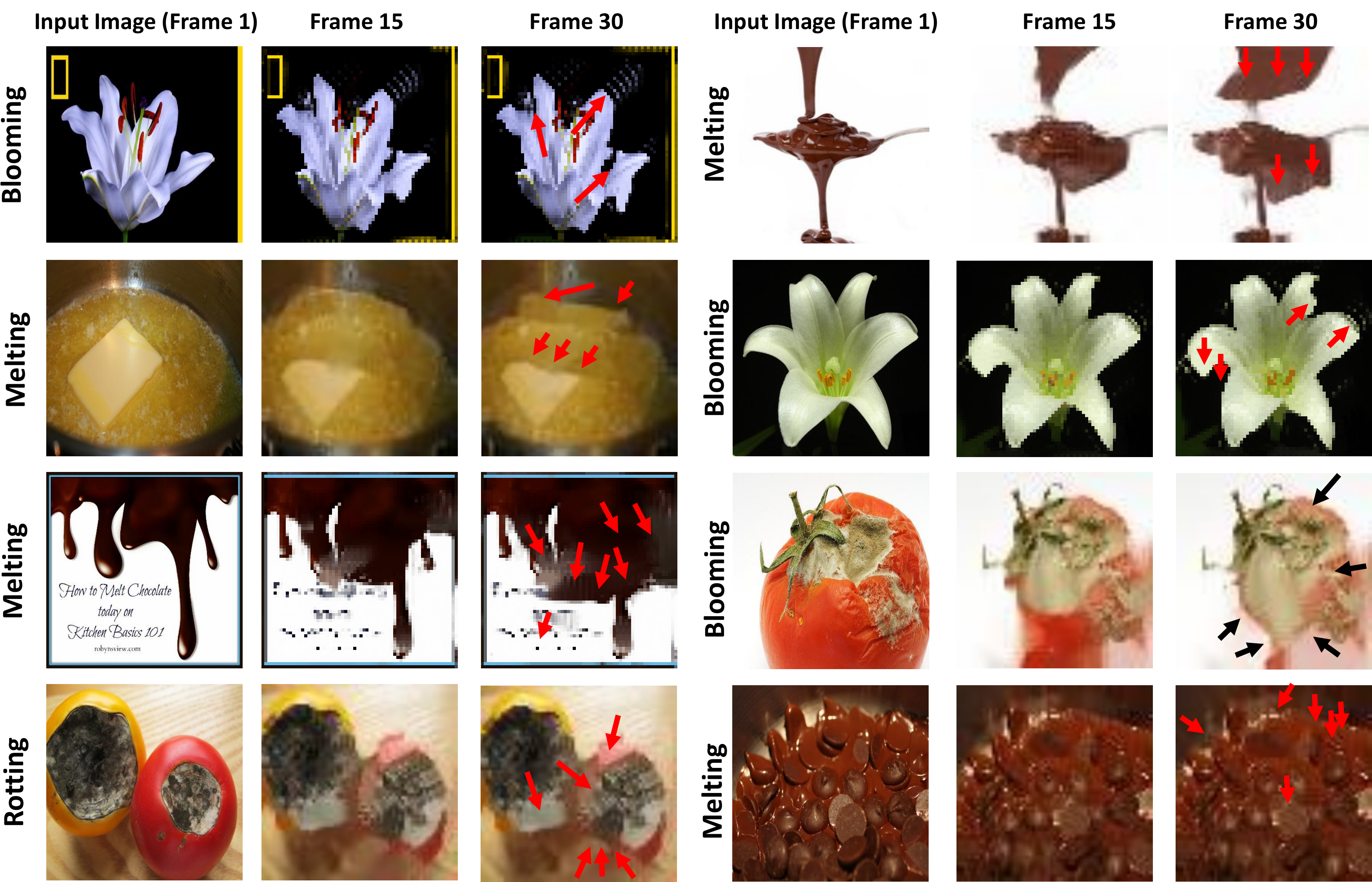}
\caption{Qualitative spatio-temporal transfer results. Notice how the petals of the flowers expand, curl and grow; how the block of butter shrinks in a realistic fashion; how chocolate expands, becomes smoother, and drips down as it melts, and how rotting fruits deform and dwindle. We encourage the reader to view our supplementary material for animations and additional examples, including transformations on atypical objects.}
\vspace{-0.5cm}
\label{fig:generation_examples}
\end{figure*}

\begin{figure}[t]
\centering
\includegraphics[width=1\linewidth]{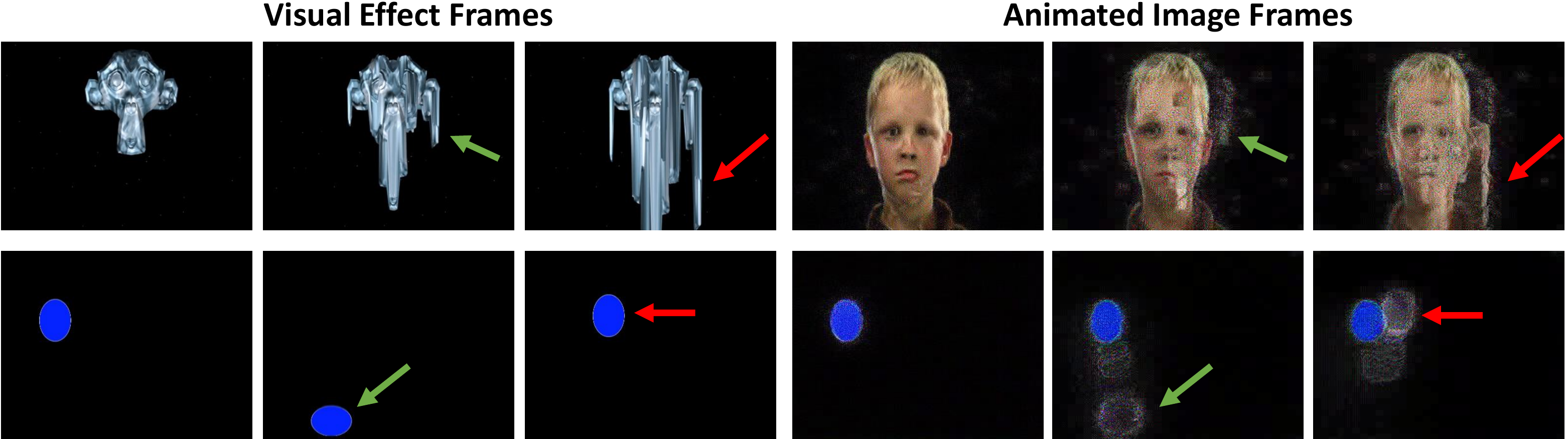}
\caption{Performing transfer by directly producing RGB values without our pixel transformer (right) compared to the ground-truth video frames (left). We observe unwanted  unnatural artifacts leaking into the synthesized frames.}
\label{fig:no_transformer}
\vspace{-1em}
\end{figure}

\subsection{Losses Evaluated}
\label{sec:baselines}

We evaluate five losses adopted from prior work. However, note that only the first one has been used in a pixel transformer framework similar to ours. The losses are: \textbf{Mean Squared Error (MSE):} We train $f(\cdot)$ using only a pixel-wise $l_2$ loss, similar to Finn~et~al.~\cite{finn2016unsupervised}. %
\textbf{MSE + Adversarial Loss (MSE+GAN)}: We train using the same loss as before, except we also use an image appearance discriminator following Zhou~et~al.\cite{zhou2016learning}.
\textbf{Adversarial Losses (GAN):} We train $f(\cdot)$ using both an optical flow and an appearance discriminator similar to Liang \cite{Liang_2017_ICCV} and Vondrick \cite{Vondrick_2017_CVPR}.
\textbf{Content + Style (C+S):} We use both the content and style loss terms from Gatys~et~al.~\cite{gatys2016image}.
\textbf{Optical Flow + Style (OF+S):} We use a Gram-matrix-based optical flow term and our style loss, similar in spirit to Tesfaldet et al. \cite{tesfaldet2017} due to its use of optical flow and style for dynamic texture generation. %

\subsection{Experimental Setup}

We evaluate our method using the timelapse video dataset of Zhou et al. ~\cite{zhou2016learning} that features four transformation categories: baking, blooming, melting, and rotting. We use a 80\%/10\%/10\% train/val/test split. Because the dataset consists of variable-length videos showing objects undergoing changes at different rates, we randomly sample frame sequences, varying both the starting frame and the skip between frames. We use a sequence length of 8 frames, which we found was enough to capture the effects. We augment the data with vertical and horizontal flips. We train our model on color frames of size $64 \times 64$ following recent video generation and prediction work \cite{videoflow, Vondrick_2017_CVPR,vondrick2016generating,finn2016unsupervised}.

\begin{table*}[t]
\centering
\resizebox{0.95\linewidth}{!}{
\begin{tabular}{cl|c|c|c|l|c|l|c|l|c|}%
\cline{3-5} \cline{7-7} \cline{9-9} \cline{11-11} 
& \multicolumn{1}{c|}{} & \multicolumn{3}{c|}{\textbf{Reconstruction}}       &  & \textbf{Human Prefs}                      &                                                 & \textbf{Human Ratings}   &  & \textbf{Data Utility}    \\ 
                                            
                                          \cline{1-1} \cline{3-5} \cline{7-7} \cline{9-9} \cline{11-11} 
\multicolumn{1}{|c|}{\textbf{Train Method}} &  & \textbf{MSE}    & \textbf{PSNR}  & \textbf{SSIM}   &  & \textbf{Percent Best}                           &                                                 & \textbf{Rating}          &  & \textbf{Accuracy}   \\                                                 %
\cline{1-1} \cline{3-5} \cline{7-7} \cline{9-9} \cline{11-11}  
\multicolumn{1}{|c|}{\textbf{First Frame}}  & \multicolumn{1}{c|}{} & \textbf{0.0039} & \textbf{30.86} & \textbf{0.9247} &  & \cellcolor[HTML]{C0C0C0}{\color[HTML]{C0C0C0} } & \cellcolor[HTML]{FFFFFF}{\color[HTML]{FFFFFF} } & \cellcolor[HTML]{C0C0C0} &  & \multicolumn{1}{l|}{\cellcolor[HTML]{C0C0C0}{\color[HTML]{FFFFFF} }}  \\ 
 \cline{1-1} \cline{3-5} \cline{7-7} \cline{9-9} \cline{11-11} 
\multicolumn{1}{|c|}{\textbf{MSE} \cite{finn2016unsupervised}}          & \multicolumn{1}{c|}{} & 0.0122          & 20.79          & 0.7283          &  & 0.1413                                          &                                                 & 2.66   ($\pm 1.16$)                  &  & 32.87\%                                                         \\ %
\cline{1-1} \cline{3-5} \cline{7-7} \cline{9-9} \cline{11-11} 
\multicolumn{1}{|c|}{\textbf{MSE+GAN} \cite{zhou2016learning}}      &                       & \textbf{0.0044} & \textbf{28.95} & \textbf{0.9170} &  & 0.0575                                          &                                                 & 3.0   ($\pm 1.04$)                   &  & 15.75\%                                                     \\ %
\cline{1-1} \cline{3-5} \cline{7-7} \cline{9-9} \cline{11-11} 
\multicolumn{1}{|c|}{\textbf{GAN} \cite{Liang_2017_ICCV}}          &                       & 0.0114          & 22.10          & 0.7510          &  & 0.0525                                          &                                                 & 2.74   ($\pm 1.05$)                  &  & 18.49\%                                                           \\ %
\cline{1-1} \cline{3-5} \cline{7-7} \cline{9-9} \cline{11-11} 
\multicolumn{1}{|c|}{\textbf{C+S} \cite{gatys2016image}}          &                       & 0.0117          & 21.46          & 0.7874          &  & \emph{0.3063}                                          &                                                 & \textbf{3.26} ($\pm 1.15$)            &  & \textbf{49.31\%}                                                  \\ %
\cline{1-1} \cline{3-5} \cline{7-7} \cline{9-9} \cline{11-11} 
\multicolumn{1}{|c|}{\textbf{OF+S} \cite{tesfaldet2017}}         &                       & \emph{0.0073}          & \emph{24.23}          & \emph{0.8671}          &  & \textbf{0.4425}                                 &                                                 & \emph{3.12} ($\pm 1.16$)                     &  & \emph{43.15}\%                                                           \\ %
\cline{1-1} \cline{3-5} \cline{7-7} \cline{9-9}  \cline{11-11} 
\end{tabular}
}
\caption{Quantitative results. Lower is better for the MSE reconstruction metric and higher is better for all other metrics. We show average results computed over all categories for each method and metric. The \textbf{First Frame} baseline simply creates a video where all frames are equal to the input first frame. We show the best result for each column in \textbf{bold}, including the first frame baseline, which obtains good reconstruction scores but is not a meaningful visual effect transfer method. The second-best result is in \emph{italics}. \textbf{OF+S} is the best method overall: it is chosen by most people as producing the highest-quality videos, and achieves reasonable reconstruction error.}
\label{table:quantitative}
\end{table*}

\subsection{Qualitative Results}
\noindent \textbf{Example transfers:} 
Fig.~\ref{fig:generation_examples} shows several example transfers for our top performing method, \textbf{OF+S}. The red arrows mark the realistic motions that this model captured. Please see supp for more video results.

\noindent \textbf{Generating without transformation:} Fig.~\ref{fig:no_transformer} contains an example of generating without using a transformation module (right, while the left shows ground-truth source video frames). We observe that pixels from the source video emerge in the black background, indicating content leaking from the source video into the output. Because our method can only perform transformations of the input pixels, it does not suffer from this problem.

\noindent \textbf{Atypical objects:} 
Because we wanted to see how well our method worked on a wider variety of objects and scenes than found in our training dataset, we assembled two datasets of still images obtained from the Internet. One dataset shows objects ordinarily found undergoing each of our visual effects, i.e.\ images for rotting may show fruit. Our human study and Fig.\ \ref{fig:generation_examples} use this dataset. The other dataset features a set of many random objects. This dataset is useful for evaluating the quality of atypical transfers (i.e.\ a deer blooming). Please view supp for these results.

\subsection{Quantitative Results}
\label{sec:quant}

We present our quantitative results in Table \ref{table:quantitative}.
Our reconstruction metrics show the distance between predicted clips on the test set and the ground truth. We emphasize that we are not interested in producing the ``correct'' spatio-temporal motion transfer result, but rather producing results that are pleasing to human observers. We observe that \textbf{First Frame}, the simple replication of a single image as all frames of a video, achieves the best reconstruction results, which indicates both that this visual effect transfer task is very challenging, and that the reconstruction metrics are not sufficient in isolation to judge quality. An MSE loss,  \textsc{MSE+GAN}, achieves the best result among the actual transfer methods; this is expected since this loss directly optimizes for low reconstruction error. The \textbf{OS+S} loss is second-best, indicating that optical flow accurately captures the motion patterns in the source videos. In addition, the use of the additional style loss ensures the natural look of the generated videos.

We next examine the human evaluation of the generated results. We observe that our human annotators preferred the \textbf{OF+S}-trained model by a large margin (44\% improvement over \textbf{C+S}). Interestingly, our evaluators gave slightly higher Likert-scale scores to their chosen \textbf{C+S} animations, but the difference between \textbf{OF+S} and \textbf{C+S} in terms of this score is within the margin of error.
We show generated videos for the best method in Fig.~\ref{fig:generation_examples}.

Finally, we analyze the data utility of each loss, in terms of producing synthetic data that can be used to train classifiers for each visual effect. We note that \textbf{C+S} outperforms \textbf{OF+S} by a small margin. This might be because the LSTM classification network extracts content features from each frame, hence a content-based loss has an advantage. 

These results highlight the differences between our proposed framework and prior work. While our pixel transformer is inspired by \cite{finn2016unsupervised}, we show that using the original loss proposed in their paper, MSE, produces low-quality visual results, as judged by our human evaluators. 
We observe that the motion-and-appearance GAN-based models representative of \cite{Liang_2017_ICCV,Vondrick_2017_CVPR} do not work well in our setting. 
Conceptually, while they learn motion dynamics, they are not concerned with transferring these to images with out-of-domain appearance. 
In contrast to \cite{tesfaldet2017} which also use an optical flow loss, here our loss is employed in the context of pixel \emph{transformations}, which allows our method to perform object deformations, rather than animate textures. 
Thus, \cite{tesfaldet2017}'s method suffers from the same problems %
in Fig.~\ref{fig:no_transformer}.
In supp, we show a comparison of our use of the optical flow loss, vs the same loss using the framework of \cite{tesfaldet2017}.

\section{Conclusion}
\label{sec:conclusion}
We studied the challenging problem of distilling spatio-temporal visual effects from videos and applying them to still images. We made progress on two fundamental challenges. First, we explored a number of image representations and loss functions for capturing and transferring visual effects. Second, we addressed the problem of unwanted content transfers by proposing a transformation-based model which is only permitted to move pixels from the input image around the output frames. Our quantitative experiments and qualitative visualizations suggest that our proposed method is able to accurately distill visual effects from videos and then apply them to images. Future work will explore how human supervision and customization options may be integrated into the transformation process for greater control on the output.

{\small
\bibliographystyle{ieee_fullname}
\bibliography{references}
}

\end{document}